\documentclass[10pt,twocolumn,letterpaper]{article}

\usepackage{cvpr}              

\usepackage{graphicx}
\usepackage{amsmath}
\usepackage{amssymb}
\usepackage{booktabs}
\usepackage{multirow}
\usepackage{color}
\usepackage{lipsum}

\usepackage[accsupp]{axessibility}  
\usepackage{algorithm}
\usepackage{algorithmic}

\usepackage[pagebackref,breaklinks,colorlinks]{hyperref}

\usepackage[capitalize]{cleveref}
\crefname{section}{Sec.}{Secs.}
\Crefname{section}{Section}{Sections}
\Crefname{table}{Table}{Tables}
\crefname{table}{Tab.}{Tabs.}

\def\cvprPaperID{7832} 
\def\confName{CVPR}
\def\confYear{2022}

\begin{document}

\title{Category-Aware Transformer Network for Better Human-Object Interaction Detection}
\vspace{-0.3CM}
\author{
Leizhen Dong, Zhimin Li\footnotemark[1], Kunlun Xu, Zhijun Zhang, Luxin Yan, Sheng Zhong, Xu Zou\footnotemark[1]
\vspace{0.2CM}\\
National Key Laboratory of Science and Technology on Multispectral Information Processing,\\
School of Artificial Intelligence and Automation, Huazhong University of Science and Technology, China\\
\vspace{-0.5cm}
{\tt\small \{dongleizhen36, lizm, xukunlun, zoux\}@hust.edu.cn}
\\
}
\maketitle
\footnotetext[1]{~Corresponding author.}
\begin{abstract}
Human-Object Interactions (HOI) detection, which aims to localize a human and a relevant object while recognizing their interaction, is crucial for understanding a still image.
Recently, tranformer-based models have significantly advanced the progress of HOI detection.
However, the capability of these models has not been fully explored since the Object Query of the model is always simply initialized as just zeros, which would affect the performance.
In this paper, we try to study the issue of promoting transformer-based HOI detectors by initializing the Object Query with category-aware semantic information.
To this end, we innovatively propose the Category-Aware Transformer Network~(CATN).
Specifically, the Object Query would be initialized via category priors represented by an external object detection model to yield a better performance.
Moreover, such category priors can be further used for enhancing the representation ability of features via the attention mechanism.
We have firstly verified our idea via the Oracle experiment by initializing the Object Query with the groundtruth category information.
And then extensive experiments have been conducted to show that a HOI detection model equipped with our idea outperforms the baseline by a large margin to achieve a new state-of-the-art result.
\end{abstract}

\begin{figure}[t]
    \centering
    \includegraphics[width=0.95\columnwidth]{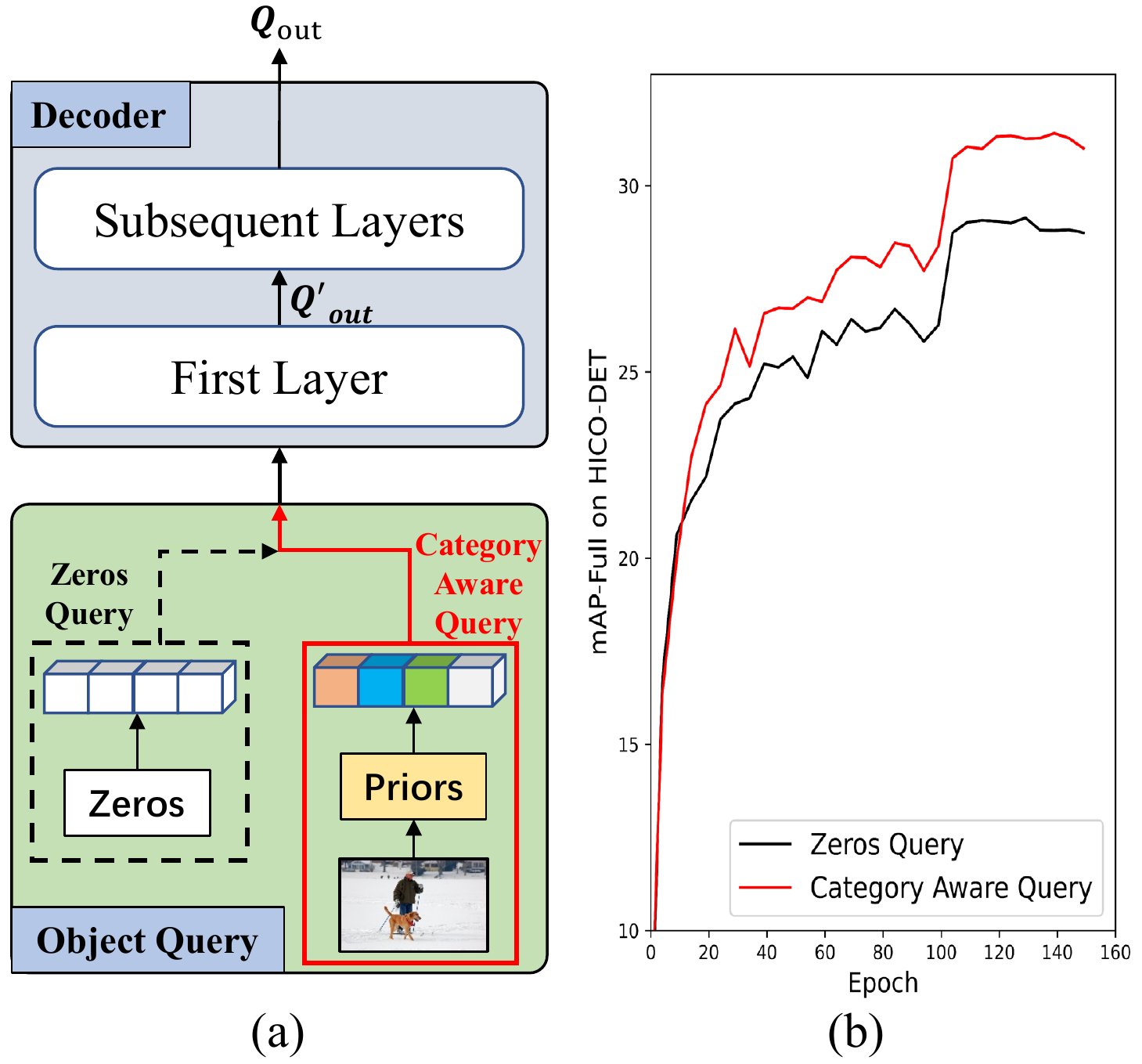} 
    \caption{
    \underline{\textbf{Black-Dotted-Box} in (a) and \textbf{Black-Curve} in (b)}: The Object Query 
    of the first decoder layer is simply initialized as just zeros in all previous transformer-based HOI detection methods. \underline{\textbf{\textcolor{red}{Red-Box}} in (a) and \textbf{\textcolor{red}{Red-Curve}} in (b)}: Our idea is to take the category-aware information as the Object Query initialization, where the training curve indicates that our method significantly promotes mAP from 29.07 to 31.03 on the HICO-DET dataset.
    }
    \label{fig1-curve}
    \vspace{-0.3cm}
\end{figure}
\vspace{-0.2cm}
\section{Introduction}
\label{sec:intro}

Human-Object Interaction~(HOI) detection, serving as a fundamental task for high-level computer vision tasks, e.g. image captioning, visual grounding, visual question answer, etc., has attracted enormous attention in recent years.
Given an image, HOI detection aims to localize the pair of human and object instances and recognize the interaction between them. A human-object interaction could be defined as the $<$human, object, verb$>$ triplet.

\par
Many two- or one-stage methods~\cite{hico-det2018learning-chao, ican2018gao, gpnn2018qi, transferable2019li, rpnn2019zhou, pastanet2020li, drg2020gao, ppdm2020liao, ipnet2020learning-wang, uniondet2020kim, atl2021houaffordance, xukunlun2022effective} have significantly advanced the process of HOI detection, while transformer-based methods~\cite{hotr2021kim, as-net2021reformulating, hoi-transformer2021, qpic2021tamura, phrse-transformer2021li, cdn2021zhang} have been remarkably proposed recently and achieved the new state-of-the-art result.

\par
Thanks to the self-attention and cross-attention mechanisms, Transformer~\cite{transformer2017attention-vaswani} has a better capability of capturing long-range dependence between different instances, which is especially suitable for the HOI detection.
HOTR~\cite{hotr2021kim} and AS-Net~\cite{as-net2021reformulating} utilize two parallel branches with transformer-decoders for performing instance detection and interaction classification respectively.
Motivated by DETR~\cite{detr2020carion}, HOI-Transformer~\cite{hoi-transformer2021} and QPIC~\cite{qpic2021tamura} adopt one transformer-decoder with several sequential layers and automatically group the different types of predictions from one query into an HOI triplet in an end-to-end manner.
\par
Though these transformer-based methods have greatly promoted the community by improving the performance and efficiency without complex grouping strategies,
there is a common issue with the Object Query\footnotemark[2] regardless of the differences in these methods.
\footnotetext[2]{The Object Query is one input of the transformer-decoder and contains $N_q$ object queries without query positional embedding in this paper.}
Specifically, the Object Query of the first decoder layer of these methods is always simply initialized as zeros since there is no previous layer for feeding semantic features~(shown as Black-Dotted-Box in Figure~\ref{fig1-curve}(a)). The capability of these models has not been fully explored due to the simple initialization of the Object Query, which would affect the performance.
Meanwhile, multi-modal information, including spatial~\cite{hico-det2018learning-chao}, posture~\cite{pairwise2018fang}, and language~\cite{pdnet2021zhong}, has been indicated to be beneficial for two-stage HOI detection models. Thus, one question remains: \textbf{how semantic information promotes a transformer-based HOI detection model?}
In this paper, we try to study the issue of elevating transformer-based HOI detectors by initializing the Object Query with category-aware semantic information.

\par
To this end, we present the Category-Aware Transformer Network~(CATN), consisting of two modules: the Category Aware Module~(CAM) and the Category-Level Attention Module~(CLAM). CAM can obtain category priors which is then applied to the initialization of the Object Query. Specifically, we use an external object detector and design a select strategy to get the categories contained in the image. After that, these categories would be transferred to corresponding word embeddings as final category priors. Moreover, these priors could be further used for enhancing the representation ability of features via the proposed CLAM.
\par
We first show that category-aware semantic information can indeed promote a transformer-based HOI detection model by the Oracle Experiment where the category priors are generated from the ground truth. Then we evaluate the effectiveness of our proposed CATN on two widely used HOI benchmarks: HICO-DET and V-COCO datasets. The contributions of our work could be summarized as:

\begin{itemize}
\item We reveal that a transformer-based HOI model can be further improved by taking category-aware semantic information as the initialization of the Object Query to achieve a new state-of-the-art result.
\item We present the Category-Aware Transformer Network (CATN), which obtains two modules: CAM for generating category priors of an image and CLAM for enhancing the representation ability of the features.
\item Extensive experiments, involving discussions of different initialization types and where to leverage the semantic information, have been conducted to demonstrate the effectiveness of the proposed idea.
\end{itemize}

\begin{figure*}[htb]
\centering
\includegraphics[width=2.0\columnwidth]{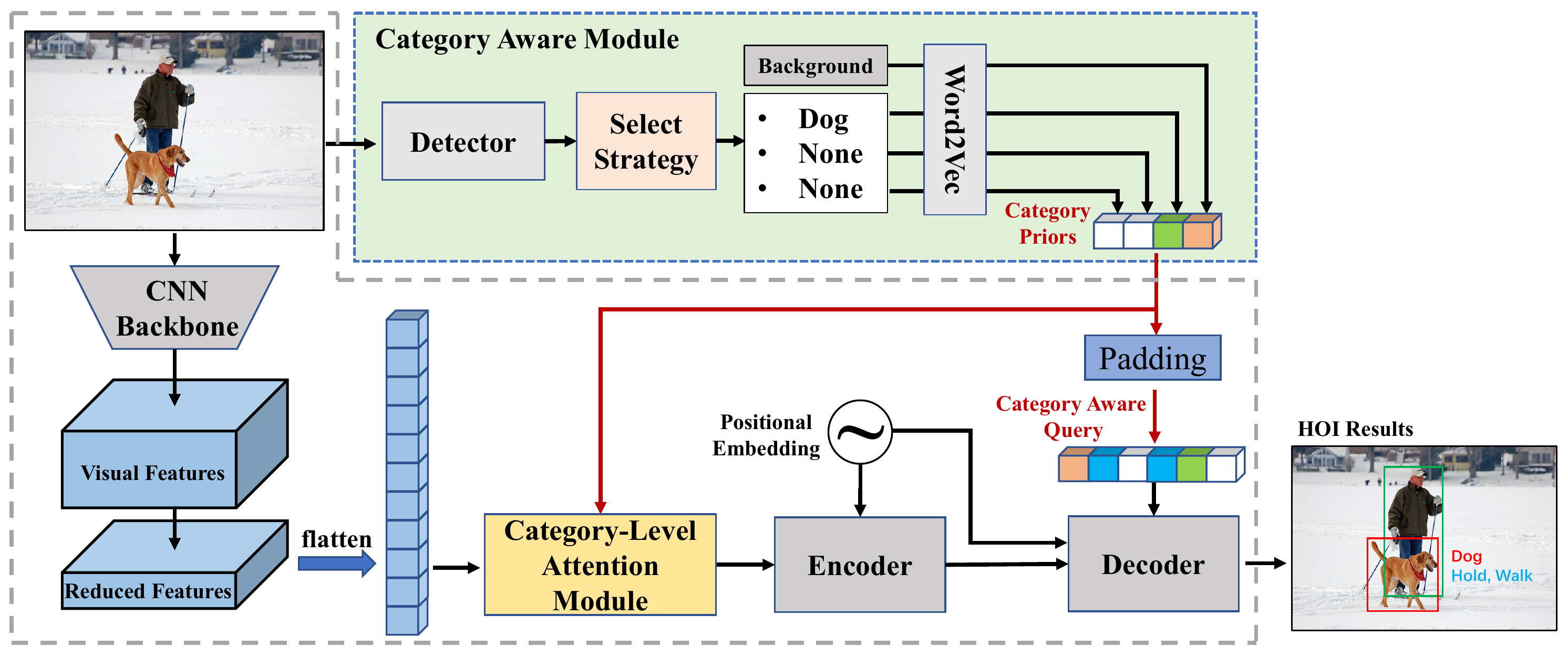} 
\caption{Overall architecture of our proposed CATN. Compared with the previous, our method contains two main components: Category Aware Module~(CAM) and Category-Level Attention Module~(CLAM).  We propose CAM which uses an external object detector to obtain category priors of the image and the priors are then applied for initializing the Object Query. Moreover, such priors can be further used in the CLAM for enhancing  the representation ability of features.}
\label{fig2-framework}
\end{figure*}
\vspace{-0.5cm}

\section{Related Works}
\label{sec:formatting}
Many remarkable methods have advanced the progress of HOI detection, which could be simply categorized into Two-, one-stage methods, and transformer-based methods.
\par
\textbf{Two-stage methods.}
Two-stage methods usually utilize a pre-trained object detector to generate human and object proposals in the first stage and then adopt an independent module to infer the multi-label interactions of each human-object pair in the second stage.
HO-RCNN~\cite{hico-det2018learning-chao} firstly presents a multi-stream architecture.
iCAN~\cite{ican2018gao} proposes an Instance-Centric Attention to aggregate the context feature of humans and objects.
In order to obtain accurate interactions, some extra information, e.g. human posture~\cite{pmfnet2019wan, pastanet2020li} and language knowledge~\cite{pdnet2021zhong}, have been introduced into HOI detection.
To better model the spatial relationship between the human and object, some GNN-based methods~\cite{gpnn2018qi, vsgnet2020ulutan, drg2020gao, rpnn2019zhou} are sequentially proposed and improve the performance.
Two-stage methods generally suffer from inefficiency due to the separate architecture, where all possible pairs of human-object proposals are predicted one after the other and the cropped features generated from the object detector maybe not suitable for interaction classification in the second stage.

\par
\textbf{One-stage Methods.}
One-stage methods are proposed to deal with the problems of high computational cost and feature mismatching appearing in two-stage methods. 
PPDM~\cite{ppdm2020liao} and IPNet~\cite{ipnet2020learning-wang} address the task of HOI as a key-point detection problem by regarding the interaction point as the mid of human-object centers. Based on the feature at the midpoint, the interactions between the human and object are predicted in a one-stage manner.
Meanwhile, UnionNet~\cite{uniondet2020kim} provides another alteration to perform HOI detection in a one-stage manner, which treats the union box of human and object bounding-box as the region of each HOI triplet. UnionNet conduct an extra branch to predict the union box and group the final HOI triplet based on IoUs.
Despite great improvement in efficiency, the performance of existing one-stage methods is limited by complex hand-crafted grouping strategies to group object detection results and the interaction predictions into final HOI triplets.

\textbf{Transformer-based Methods.}
Recently, transformer~\cite{transformer2017attention-vaswani}, with a good capability of capturing the long-range dependency, has been introduced to the HOI detection and brings a significant performance improvement.
HOTR~\cite{hotr2021kim} and AS-Net~\cite{as-net2021reformulating} combine the advantages of both one-stage method and transformer, and utilize two parallel decoders to predict human-object proposals and interactions respectively.
Apart from the above methods,  HOI-Transformer~\cite{hoi-transformer2021} and QPIC~\cite{qpic2021tamura} extend DETR~\cite{detr2020carion} to the HOI detection, which directly defines the predictions from a query as the   HOI triplet without the complex grouping strategy.
\par
Although significant performance is obtained by these transformer-based methods, they have a common issue that the Object Query is initialized with zeros as illustrated in Figure~\ref{fig1-curve}(a). In this paper, we study the issue of how to promote a transformer-based HOI detector by initializing the Object Query with category-aware semantic information.

\textbf{Category Information and HOI Detection.} Category Information is a kind of semantic information indeed, which represents the object categories in an image. The effectiveness of such information has been demonstrated in several domains. Different from part-of-speech category for image captioning~\cite{kim2019dense} or the category shape for 3D-Reconstruction~\cite{runz2020frodo}, we studied object category information since the instance has category-aware relation in HOI detection, e.g. person-eat-apple, person-ride-bike, etc.

\vspace{-0.5cm}
\section{Approach}
\subsection{Overview}
In this section, we present our Category-Aware Transformer Network~(CATN), trying to improve the performance of transformer-based HOI detectors with category priors.
Firstly, we start with the overall architecture of our proposed CATN.
Secondly, we detailedly introduce the Category Aware Module~(CAM) to extract category priors of an image, which then are applied to the initialization of the Object Query of the first decoder layer.
Moreover, we propose the Category-Level Attention Module~(CLAM) to enhance the capability of features with such priors.
Finally, we modify the matching cost, used in bipartite matching, for better matching between the ground-truths and $N_q$ predictions.

\subsection{CATN Architecture}
The overall pipeline of our CATN is illustrated in Figure~\ref{fig2-framework}, which is similar to previous transformer-based methods except for additional proposed CAM and CLAM.
\par
\noindent \textbf{Backbone.} Given an RGB image, we firstly adopt a CNN-based backbone to extract a visual feature map denoted as $I_c \in \mathbb{R}^{D_c\times h\times w}$. Then a convolution layer with a kernel size of $1 \times 1$ is utilized to reduce the channel dimension from $D_c$ to $D_d$, where $D_c$, $D_d$ are 2,048 and 256 by default. Then the visual feature map is flattened and denoted as $I_{visual} \in \mathbb{R}^{D_d\times hw}$. After that, we adopt the CLAM to enhance the features from CNN with category priors and denote the output feature map as $I_{CLAM} \in \mathbb{R}^{D_d\times hw}$.

\par
\noindent \textbf{Encoder.} The transformer encoder aims to improve the capability of capturing long-range dependence. It is a stack of multiple encoder layers, where each layer mainly consists of a self-attention layer and a feed-forward~(FFN) layer. To make the flatten features spatially aware, a fixed Spatial Positional Encoding, denoted as $P_S \in \mathbb{R}^{D_d \times hw}$, is conducted and fed into each encoder layer with the features.
The calculation of the transformer encoder could be expressed as:
\vspace{-0.5cm}
\begin{align}
    I_{enc} = f_{enc \times N_{enc}}(I_{CLAM}, P_{S})
\end{align}
where $ f_{enc}$ indicates the function of one encoder layer, $N_{enc}$ is the number of stacked layers, and $I_{enc} \in \mathbb{R}^{D_d \times hw}$ is the output feature and then fed into the following decoder.

\noindent \textbf{Decoder.}
The transformer decoder aims to transform a set of object queries $Q_{zeros} \in \mathbb{R}^{N_q \times D_d}$ (with query positional embedding $P_{Q} \in \mathbb{R}^{N_q \times D_d}$ whose parameters are learnable) to another set of output queries $Q_{out} \in \mathbb{R}^{N_q \times D_d}$. It is also a stack of decoder layers. Apart from selt-attention and FFN, each decoder layer contains an additional cross-attention layer, which is used to aggregate the features $I_{enc}$ output from encoder into $N_q$ queries.
\par
In our CATN, $Q_{zeros}$ is replaced with Category Aware Query~(CAQ), denoted as $Q_{CA} \in \mathbb{R}^{N_q \times D_d}$, which is generated via category priors.
The calculation of the transformer decoder could be expressed as:
\begin{align}
    Q_{out} = f_{dec \times N_{dec}}(Q_{CA}, P_{Q}, I_{enc}, P_{S})
\end{align}
where $f_{dec}$ indicates the function of one decoder layer and $N_{dec}$ is the number of stacked decoder layers.

\noindent \textbf{Prediction Head.}
In our experiments, an HOI triplet consists of four elements: the human bounding box, the object bounding box, the object category with its confidence, and multiple verb categories with their confidence. Based on the above definition, four feed-forward networks~(FFNs) are conducted on each output query as follows:

\vspace{-0.5cm}
\begin{align}
    \begin{cases}
        b_h^i = \sigma(f_{h,b}(Q_{out}^i)) \\
        b_o^i = \sigma(f_{o,b}(Q_{out}^i)) \\
        c_o^i = \varsigma(f_{o,c}(Q_{out}^i)) \\
        c_v^i = \sigma(f_{v,c}(Q_{out}^i)) \\
    \end{cases}
\end{align}
where $i$ indicates the index of outputting queries and the ground-truths, and $\sigma, \varsigma$ are the sigmoid and softmax functions respectively.

\subsection{Category Aware Module}
As mentioned above, the Object Query of the first decoder layer is simply initialized with zeros since there is no last layer where we argue this may affect the performance. In this section, we detailedly introduce the Category Aware Module~(CAM) to extract the category priors of an image which then are used for Category Aware Query and CLAM.
\par
The Blue-Dotted-Box in Figure~\ref{fig2-framework} describes the proposed CAM. Given an image, we firstly utilize an external object detector, e.g. Faster-RCNN, to perform object detection and only reserve the results with confidence scores higher than the detection threshold $T_{det}$. Since we focus on studying the effect of category-aware semantic information on HOI detectors and avoid the influence of other factors, we directly discard the bounding box of each prediction and only utilize the category with its confidence score.
\par
\noindent \textbf{Select Strategy.}
Based on their categories, the rest results can be divided into different sets $\Omega = \{ \Omega_1, \Omega_2, ..., \Omega_K \}$, where $K$ is the total number of categories in the dataset and $\Omega_i$ represents a set of detection results whose category is the i-th category denoted as $c_i$. After that, we calculate the confidence score as follows:
\begin{align}
    S_{c_i} =
    \begin{cases}
        max(\Omega_i) + \frac{|\Omega_i|}{2} \times mean(\Omega_i) , &|\Omega_i| \neq 0\\
        0 , &|\Omega_i| = 0\\
    \end{cases}
\end{align}
where $S_{c_i}$ represents the probability of category $c_i$ contained in the image and $|.|$ indicates the number of the set.
\par
With these statistics, we firstly select a threshold $T_{can}$ for a set of candidate categories $\Omega_{can} = \{c_i | \sum_{i = 1}^{K} S_{c_i} \geq T_{can}\}$ and re-rank them based on $S_{c_i}$. Then $Top^{(N_c - 1)}$ categories from $\Omega_{can}$ with a fixed category (named as `background') are set as the prior categories of an image, where the `background' is used as the placeholder for matching if no relevant instance is obtained by the detector in CAM and the detail is discussed in Section 3.5. Note that the rest category will be filled with `None' if the number of categories in $\Omega_{can}$ is lower than $N_{c} - 1$. We denote the final prior categories of the image as $C^* = \{ c_i | c_i \in C_{can} \cup None\}_i^{N_{c}}$.
\par
\noindent \textbf{Category Priors.}
We transform prior categories of an image to the word embedding vectors which could be used in the following module. To this end, we utilize a pre-trained word2vector model, e.g. fastText~\cite{fastText2017mikolov-advances}, to generate the category priors of an image.

\vspace{-0.5cm}
\begin{align}
    E_{prior} = \{f_{FC}(f_{w2v}(c_i))  | c_i \in C^*\}
\end{align}
where $f_{w2v}(c_i)$ is to obtain the embedding vector of $c_i$ category and $f_{FC}$ is a fully connected layer to adjust the dimension of the embedding. Especially, the embeddings of all object categories are calculated beforehand and saved locally. Regardless of training or inference, there is only a slight increase in computation cost due to the fully connected layer. In addition, we also evaluate several different word2vector models and the experimental results are shown in the later section.

\par
\noindent \textbf{Category Aware Query.}
An image may contain more than one HOI triplet with the same category and the number of prior categories $N_c$ is usually much smaller than the number of queries $N_q$.
Thus, we generate $Q_{CA} \in R^{N_q \times D_d}$  by simply repeating the $E_{prior}$ vectors $\frac{N_q}{N_c}$ times as follows.

\vspace{-0.5cm}
\begin{align}
Q_{CA} = Repeat(E_{prior}, N_q, N_{c})
\end{align}
Finally, we use $Q_{CA}$ as initial values of the Object Query. 

\begin{figure}[t]
    \centering
    \includegraphics[width=0.75\columnwidth]{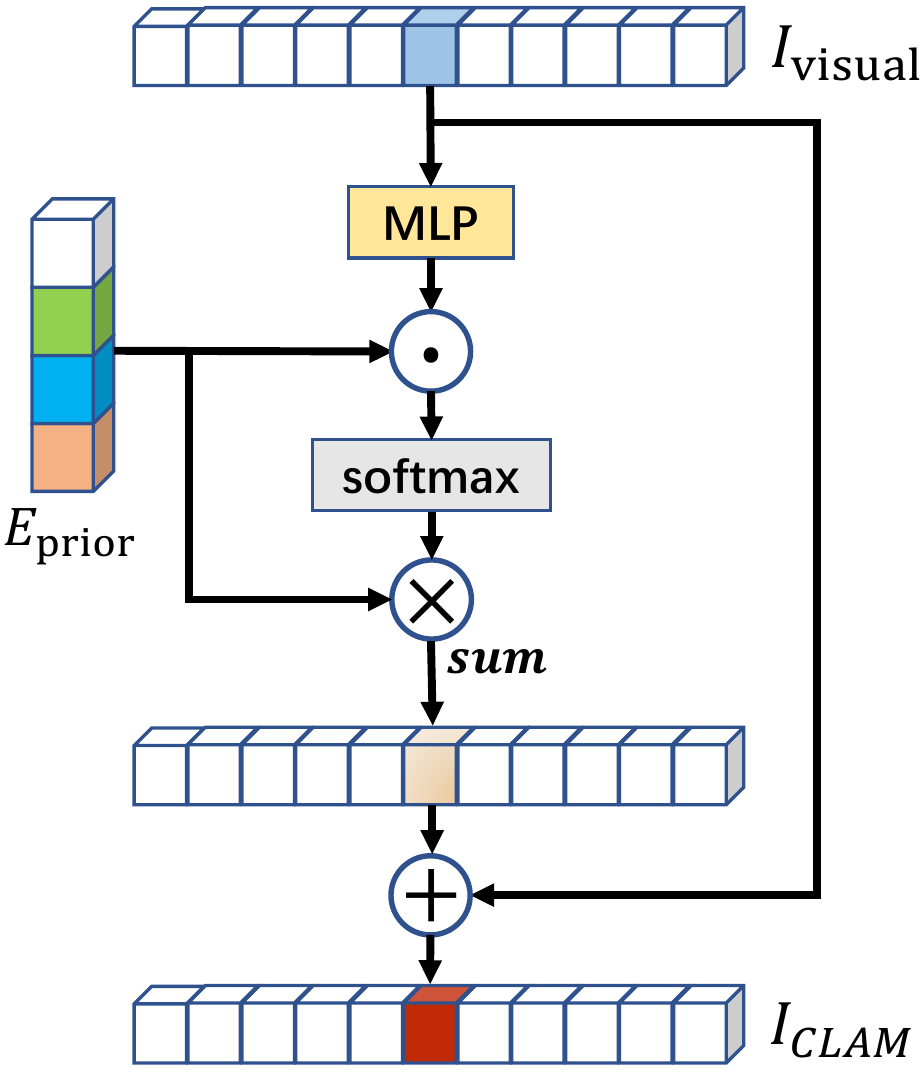}
    \caption{The pipeline of our proposed Category-Level Attention Module~(CLAM). Each cuboid indicates a feature vector with the shape of $1 \times D_d$. $\cdot$, $\times$, + mean dot-product, multiplication and element-wise addition respectively.}
    \label{fig:clam}
    \vspace{-0.2cm}
\end{figure}

\subsection{Category-Level Attention Module}
For maximizing the capability of category information, we also propose an attention mechanism, named as Category-Level Attention Module~(CLAM), to enhance the representation ability of features output from backbone.
As illustrated in Figure~\ref{fig:clam}, to clearly describe the entire workflow, we take the visual feature in one location as an example of this module and denote the feature as $X_{visual} \in \mathbb{R}^{1 \times D_d}$, while features work consistently.
\par
The visual feature $X_{visual}$ is firstly projected to another $D_d$-dimensional vector, denoted as $\hat{X}_{visual}\in \mathbb{R}^{1 \times D_d}$, via a Muti-Layer Perception~(MLP). The MLP contains an FC layer without BatchNorm and ReLU and is used to transform the feature from visual space to word space. Then we measure the similarity between the vector and category embedding by dot product operation, and a softmax function is followed to normalize the similarity values of all categories.

\vspace{-0.5cm}
\begin{align}
    W_{att} = Softmax(MLP(X_{visual}) \cdot E_{prior}^T)
\end{align}
where $W_{att} \in \mathbb{R}^{1 \times N_{c}}$ represents the attention weights of prior categories on the feature and `$\cdot$' means Dot-Product.
Specifically, the weight in a category is high if the feature contains rich information related to the category.
To make the features category aware, we aggregate all word embeddings of prior categories with corresponding weights into one vector $X_{word} \in \mathbb{R}^{1 \times D_d}$ and then add the vector back to the original visual feature $X_{visual}$. The feature aggregation can be written as:

\vspace{-0.5cm}
\begin{align}
    X_{clam} = X_{app} + \sum_{j=1}^{N_c} w_{att}^j \times E_{prior}^j
\end{align}
where $w_{att}^{j}$ is a value and indicates the attention weight of j-th category belonging to the prior categories, $E_{prior}^{j}$ is the embedding of j-th category, `+' represents element-wise addition, and `$\times$' mean the multiplication between the scalar $w_{att}^j$ and each element of the vector $E_{prior}^j$ respectively.

\par
Our proposed CLAM has several advantages. Compared to instance-level attention mechanism~\cite{ican2018gao}, ours is category-level and has lower requirements on the accuracy of bounding boxes. Meanwhile, we add the weighted average word embedding with category information to the originally visual features. Therefore, the aggregated features, output from our CLAN, not only have the capability of aware visual information but also are aware of category information.

\vspace{-0.5cm}
\subsection{Matching Cost \& Training Loss}
For training transformer-based models~\cite{detr2020carion} with a set of prediction results, the bipartite-matching algorithm is publicly used to automatically match a ground-truth with at most one prediction, which would suppress the problem of redundant predictions. To this end, two types of losses are introduced below.
\par
\noindent \textbf{Modified Matching Cost.} Matching cost is conducted to measure the similarity between the ground truth and an HOI prediction and assign the label of each query whether a positive or a negative. Firstly, we calculate the matching cost $H \in \mathbb{R}^{N_{gt} \times N_q}$ following Formula 1 in~\cite{qpic2021tamura}, where $N_{gt}$ is the number of ground-truth HOI triplet and $H_{i, j}$ indicates the matching cost between i-th ground-truth and j-th prediction generated from j-th output query. Then, we modify the matching cost by $\hat{H}_{i,j}= H_{i,j} + Cost_{i,j}$ and the external cost $Cost_{i, j}$ is defined as follows:

\vspace{-0.3cm}
\begin{align}
    Cost_{i, j}=
    \begin{cases}
        0 , &C(q_{j})=C(GT_i)\\
        v , &C(q_{j})=``Background''\\
        2v , &C(q_{j})=``None''\\
        2v , &Else
    \end{cases}
\end{align}
where $C$ represents the corresponding object category, $q_j$ is the j-th query of $Q_{CA}$, $GT_i$ means the i-th ground-truth triplet in the image. The $Cost_{i,j}$ is used to make the matching cost $H_{i, j}$ higher when the object categories of j-th query and i-th ground-truth are different. Meanwhile, the experimental results show that there is no difference when the value is higher than a threshold. Thus we empirically set $v$ as 500.
Finally, we utilize the Hungarian Algorithm~\cite{hungarian1955kuhn} to perform the optimal assignment $\hat{\omega} = argmin_{\omega \in \Omega_{N_q}} \sum_{i=1}^{N_q}\hat{H}_{i,\omega(i)}$, where only $N_{gt}$ predictions in $\hat{\omega}$ are set as positive and the rest are negative.
\par
With the above modifications, a ground-truth will match the query where their object category are the same. In addition, if the object category of a ground truth is not included in prior categories, the ground truth will preferentially match the query whose prior category is ``background''. Thus the modified cost shrinks the matching space between the ground truth and the predictions.
\par
\noindent \textbf{Training Loss.} Based on the above label assignment, the training loss is calculated to optimize the parameters of our CATN model. We directly adopt equations 6$\sim$10 in~\cite{qpic2021tamura} and keep the weights consistent, which reveals that the performance improvement is obtained by our proposed category priors, not hyper-parameters.

\section{Experiments}

\subsection{Datasets \& Metrics}
\noindent\textbf{Datasets.} We conduct experiments on two widely used datasets to verify the effectiveness of our model.
\textbf{V-COCO}~\cite{vcoco2015visual-gupta}, a subset of COCO~\cite{coco2014microsoft-lin} dataset, consists of 2,533 training images, 2,867 valuation images, and 4,946 test images respectively. There are 16,199 human instances and each instance has a set of binary labels for 29 different actions.
\textbf{HICO-DET}~\cite{hico-det2018learning-chao} is the largest dataset in HOI detection. There are 38,118 training images and 9,658 test images respectively with totally more than 150k HOI annotations. It has 600 hoi categories~(Full) with 117 verb categories and 80 object categories, which can be further divided into 118 categories~(Rare) and 462 categories~(Non-Rare) based on the number of instances in the training set.

\par
\noindent \textbf{Evaluation Metrics.}
Following the standard rule~\cite{vcoco2015visual-gupta}, we
use the commonly used role mean average precision (mAP) to evaluate the model performance for both benchmarks.
An HOI prediction is regarded as a true positive if the categories of the object and verbs are correct, and the predicted bounding boxes of the human and object are localized accurately where the IoUs are greater than 0.5 with the corresponding ground truth.

\subsection{Implementation Details}
We conduct our experiments with the publicly available PyTorch framework~\cite{pytorch2019paszke}.
\par
For the external detector used in CAM to obtain category priors, we adopt Faster-RCNN-FPN~\cite{fasterrcnn2015ren, fpn2017lin} with ResNet-50~\cite{resnet2016he} as the backbone to perform object detection. For better performance, we use COCO pre-trained weights and then fine-tune the model on both HICO-DET and V-COCO datasets. During training, we drop the weight of regressing Bbox in loss cost from 1.0 to 0.2 and keep the other hyper-parameters consistent as default. Meanwhile, the human category is discarded since the number of humans is dominant. We set the batch size to 4 and use SGD as the optimizer with a learning rate of 0.01, a weight decay of 0.0001, and a momentum of 0.9. We train the model for 12 epochs with twice the learning rate decay at epoch 8, 11 by 10 times respectively. The detection threshold $T_{det}$ is set to 0.15. The prior threshold $T_{can}$, used for category priors, is set to 0.3 and 0.4 for HICO-DET and V-COCO respectively. The number $N_{c}$ is set to 4 and 5 for two datasets respectively. To obtain better category priors, we adopt some commonly used augmentation strategies, including random scales, random flip, color jittering, and random corp augmentation.

\par
For our CATN, ResNet-50 is used as the backbone, the number of encoder and decoder layers are both set to 6 and the number of Object Query $N_q$ is set to 100. We initialize the network with parameters of DETR~\cite{detr2020carion} pre-trained on the COCO dataset. During training, we set the batch size to 16, the backbone's learning rate to 1e-5, the transformer's learning rate to 1e-4, and weight decay to 1e-4. The model is trained for 150 epochs totally on both datasets with once learning rate decreased by 10 times at epoch 100. Following DETR, scale augmentation, scaling the input image such that the shortest side is at least 480 and at most 800 pixels while the longest at most 1333, is adopted for better performance in training.
\par
Note that the category embedding is generated by fastText~\cite{fastText2017mikolov-advances} and the ``baseline'' indicates the QPIC~\cite{qpic2021tamura} with ResNet-50~\cite{resnet2016he} if there are no additional comments.

\begin{table}[t]
    \centering
    \resizebox{0.95\columnwidth}{!}{
    \begin{tabular}{c | c | c c c}
    \hline
        Method & Query & Full & Rare & Non-Rare \\
    \hline
        baseline & Zeros & 29.07 & 21.85 & 31.23\\
        Ours & CAQ* & 37.17 & 31.65 & 38.81\\
    \hline
        \multicolumn{2}{c|}{Improvement}
        & (8.10 \textcolor{red}{$\uparrow$})
        & (9.80 \textcolor{red}{$\uparrow$})
        & (7.58 \textcolor{red}{$\uparrow$})\\
    \hline
    \end{tabular}
    }
    \caption{Oracle experiment on HICO-DET dataset. Zeros and CAQ represent that the Object Query is initialized with zero-values or category priors respectively. * indicates such category priors generated from the ground truth. The performance is tremendously promoted once category priors are adopted for initializing the Object Query. This phenomenon directly indicates the rationality of introducing category priors.
    }
    \label{tab:oracle_experiment}
    \vspace{-0.2cm}
\end{table}
\subsection{Oracle Experiment}
To verify the effectiveness of our idea that the performance of transformer-based HOI detectors could be further improved by initializing the Object Query with category-aware semantic information, we firstly conduct the oracle experiment where the category priors of an image are simply generated from the ground truth.
\par
Table~\ref{tab:oracle_experiment} illustrates the experimental results on HICO-DET. In this experiment, we select QPIC as the baseline and only apply such priors to the Object Query without the proposed CLAM. Compared with the baseline, our method achieves a great performance improvement on all three default settings. With such category priors, the `Full' performance is improved from 29.09 to 37.17  with a 27.8\% relative performance gain and especially the `Rare' performance is improved from 21.85 to 31.65 with a 44.8\% relative performance gain. This simple experiment with great performance gain verifies the effectiveness of our idea and supports subsequent detailed experiments.

\begin{table*}[t]
    \centering
    \begin{tabular}{l c c | c c c | c c c }
    \hline
    \hline
        &  &  &  \multicolumn{3}{c|}{Default} & \multicolumn{3}{c}{Known Object}\\

       Methods & Backbone & Detector  & Full & Rare & Non-Rare & Full & Rare & Non-Rare \\
    \hline
    \multicolumn{2}{l}{\textit{Two-Stage Methods}}  & & & & & &\\
        HO-RCNN~\cite{hico-det2018learning-chao} & CaffeNet & C & 7.81 & 5.37 & 8.54 & 10.41 & 8.94 & 10.85\\
        InteractNet~\cite{InteractNet2018detecting-gkioxari} & R50-FPN & C & 9.94 & 7.16 & 10.77 & - & - & -\\
        GPNN~\cite{gpnn2018qi} & R101 & C & 13.11 & 9.34 & 14.23 & - & - & -\\
        iCAN~\cite{ican2018gao} & R50 & C & 14.84 & 10.45 & 16.15 & 16.26 & 11.33 & 17.73\\
        PMFNet~\cite{pmfnet2019wan} & R50-FPN & C & 17.46 & 15.65 & 18.00 & 20.34 & 17.47 & 21.20\\
        VSGNet~\cite{vsgnet2020ulutan} & R152 & C & 19.80 & 14.63 & 20.87 & - & - & -\\
        PDNet~\cite{pdnet2021zhong} & R152 & C & 20.81 & 15.90 & 22.28 & 24.78 & 18.88 & 26.54\\
        FCMNet~\cite{fcmnet2020amplifying-liu} & R50 & C & 20.41 & 17.34 & 21.56 & 22.04 & 18.97 & 23.12\\
        PastaNet~\cite{pastanet2020li} & R50 & C & 22.65 & 21.17 & 23.09 & 24.53 & 23.00 & 24.99\\
        VCL~\cite{vcl2020hou} & R101 & H & 23.63 & 17.21 & 25.55 & 25.98 & 19.12 & 28.03\\
        DRG~\cite{drg2020gao} & R50-FPN & H & 24.53 & 19.47 & 26.04 & 27.98 & 23.11 & 29.43\\
    \hline
    \multicolumn{2}{l}{\textit{One-Stage Methods}}  & & & & & & &\\
        UnionDet~\cite{uniondet2020kim} & R50-FPN & H & 17.58 & 11.52 & 19.33 & 19.76 & 14.68 & 21.27\\
        IPNet~\cite{ipnet2020learning-wang} & HG-104 & C & 19.56 & 12.79 & 21.58 & 22.05 & 15.77 & 23.92\\
        PPDM~\cite{ppdm2020liao} & HG-104 & H & 21.73 & 13.78 & 24.10 & 24.58 & 16.65 & 26.84\\
    \hline
    \multicolumn{2}{l}{\textit{Transformer-based Methods}} & & & & & & &\\
        HOI Transformer~\cite{hoi-transformer2021} & R50 & - & 23.46 & 16.91 & 25.41 & 26.15 & 19.24 & 28.22\\
        HOTR~\cite{hotr2021kim} & R50 & - & 25.10 & 17.34 & 27.42 & - & - & - \\
        AS-Net~\cite{as-net2021reformulating} & R50 & - & 28.87 & 24.25 & 30.25 & 31.74 & \underline{27.07} & 33.14\\
        QPIC~\cite{qpic2021tamura} & R50 & - & 29.07 & 21.85 & 31.23 & 31.68 & 24.14 & 33.93\\
    \hline
    \multicolumn{2}{l}{\textit{Ours}} & & & & & & &\\
        \textbf{CATN~(with fastText~\cite{fastText2017mikolov-advances})} & R50 & H & 31.62 & 24.28 & \underline{33.79} & 33.53 & 26.53 & 35.92\\
        \textbf{CATN~(with BERT~\cite{bert2018devlin})} & R50 & H & \textbf{31.86} & \textbf{25.15} & \textbf{33.84} & \textbf{34.44} & \textbf{27.69} & \textbf{36.45}\\
        \textbf{CATN~(with CLIP~\cite{clip2021radfordlearning})} & R50 & H & \underline{31.71} & \underline{24.82} & 33.77 & \underline{33.96} & 26.37 & \underline{36.23}\\
    \hline
    \hline
    \end{tabular}
    \caption{Comparison against state-of-the-art methods on HICO-DET dataset. For Detector, C means that the detector is trained on COCO dataset, while H means that the detector is then fine-tuned on HICO-DET dataset. `Default' and `Known Object' are two evaluation modes following the standard rule. ``fastText'', ``BERT'', ``CLIP'' means that the embeddings of prior categories are obtained from these pre-trained word2vector models. The BEST and the SECOND BEST performances are highlighted in \textbf{bold} and \underline{underlined} respectively. Our proposed CATN outperforms the previous by a large margin to achieve new state-of-the-art results on both evaluation modes.}
    \label{tab:sota-hico}
\end{table*}

\begin{table}[thb]
    \centering
    \resizebox{0.9\columnwidth}{!}{
    \begin{tabular}{l c | c }
    \hline
         Methods & Backbone & AProle  \\
    \hline
         VCL~\cite{vcl2020hou} & R50-FPN & 48.3 \\
         DRG~\cite{drg2020gao} & R50-FPN & 51.0 \\
         PDNet~\cite{pdnet2021zhong} & R152 & 52.6 \\
    \hline
         UnionBox~\cite{uniondet2020kim} & R50-FPN & 47.5 \\
         IPNet~\cite{ipnet2020learning-wang} & HG-104 & 51.0 \\
    \hline
         HOI Transformer~\cite{hoi-transformer2021} & R50 & 52.9 \\
         HOTR~\cite{hotr2021kim} & R50 & 55.2 \\
         AS-Net~\cite{as-net2021reformulating} & R50 & 53.9 \\
         QPIC~\cite{qpic2021tamura} & R50 & 58.8 \\
         \textbf{CATN~(with fastText~\cite{fastText2017mikolov-advances})} & R50 & \textbf{60.1} \\
    \hline
    \end{tabular}
    }
    \caption{Comparison against state-of-the-art methods on V-COCO dataset. The BEST performances are high-lighted in \textbf{bold}. Ours also outperforms others to achieve a new state-of-the-art result.}
    \label{tab:sota-vcoco}

    \vspace{-0.5cm}
\end{table}

\subsection{Comparison to the State-of-The-Art}
In this section, we use the proposed CAM to obtain category priors of an image and compare our proposed CATN with other state-of-the-art methods on two public benchmarks.
\textbf{HICO-DET.}
To verify the effectiveness of our proposed idea, we adopt several different word2vector models including fastText\cite{fastText2017mikolov-advances}, BERT~\cite{bert2018devlin} and CLIP~\cite{clip2021radfordlearning}, to obtain the category-aware semantic information and conducts the experiments on HICO-DET dataset.
As shown in Table~\ref{tab:sota-hico}, our proposed method obtains the significant performance improvement on both ``Default'' and ``Known-Object'' evaluation modes. Especially, the experiment with BERT~\cite{bert2018devlin} has achieved the new state-of-the-art result, which promotes the mAP-full from 29.07 to 31.86 in Default mode and from 31.74 to 34.44 in Known-Object mode.
\textbf{V-COCO.}
We also evaluate our proposed CATN on V-COCO dataset. A similar performance gain is obtained as shown in Tabel~\ref{tab:sota-vcoco}. Compared with previous methods, our method also achieves a new state-of-the-art result. With the embeddings generated by fastText~\cite{fastText2017mikolov-advances}, we reach an AP-role of 60.1, which obtains 1.3 points performance gain than the second-best method.

\subsection{Ablations Study}
\textbf{The effectiveness of each component in our CATN.} In order to make a clearer study of the impact of each component on the overall performance, supplementary ablation experiments are conducted on the HICO-DET dataset. The results in Default evaluation mode are shown in Table~\ref{tab:ablation}.
Initializing the Object Query with category-aware semantic information instead of just zeros~\cite{qpic2021tamura} can effectively improve mAP from 29.07 to 30.82, which indicates the superiority of our main idea on HOI detection. Modifying the matching cost can also promote mAP to 31.03 with a gain of 0.21 mAP. Illustrated as line 4, the performance could be further improved from 31.03 to 31.62 when our proposed CLAM enhances the representation ability of features via the category-aware semantic information. Moreover, we visualize an example of the attention map in the supplementary file to demonstrate the effectiveness of our CLAM.


\begin{table}[t]
    \centering
    \begin{tabular}{c | c | c c c | c}
    \hline
          & Method & CAQ & MMC & CLAM & mAP \\
    \hline
        1 & baseline & - & - & - & 29.07\\
    \hline
        2 & \multirow{3}{*}{CATN} & \checkmark & & & 30.82\\
        3 &  & \checkmark & \checkmark & & 31.03 \\
        4 &  & \checkmark & \checkmark & \checkmark & \textbf{31.62} \\
    \hline
    \end{tabular}
    \caption{Ablation studies on the effectiveness of each module in our CATN on HICO-DET dataset. $\checkmark$ represents the component is used. ``CAQ'' means the Object Query is initialized with category-aware semantic information. ``MMC'' indicates our modified matching cost. ``CLAM'' represents the proposed Category-Level Attention Module.}
    \label{tab:ablation}
    \vspace{-0.5cm}
\end{table}

\subsection{Discussion}
\par
\begin{figure}
    \centering
    \includegraphics[width=0.95\columnwidth]{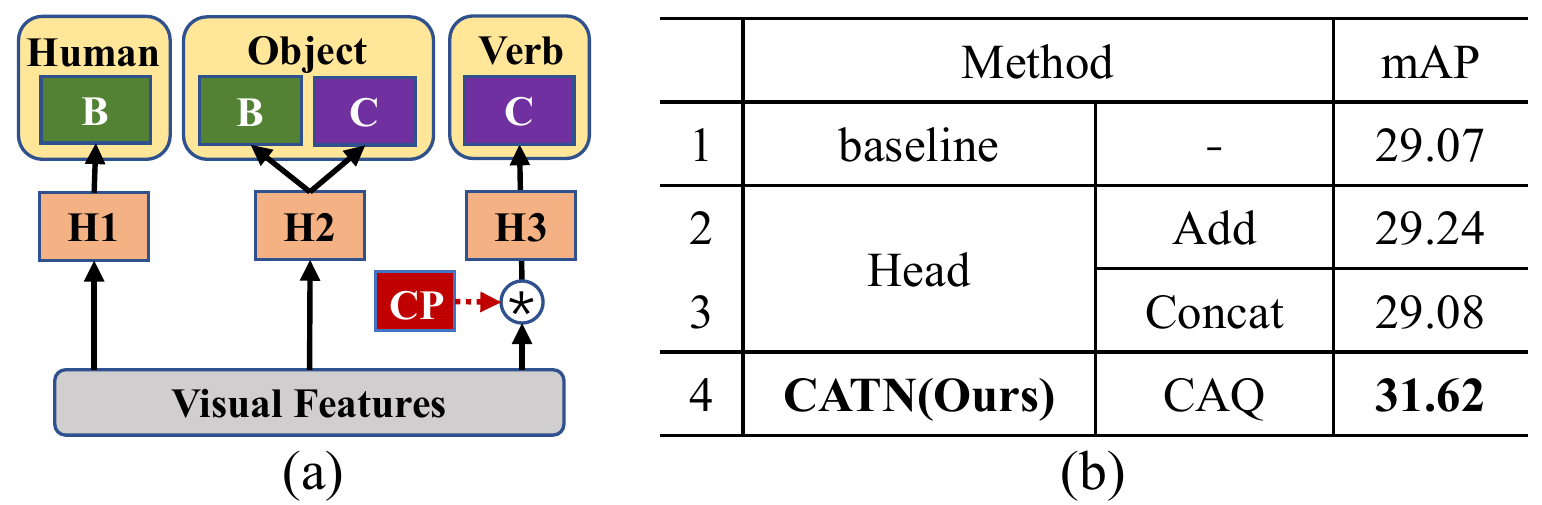}
    \caption{The impacts of where to leverage the category priors.
    ``H'', ``B'', ``C'' are prediction heads, bounding boxes, and categories respectively.
    Similar to~\cite{drg2020gao}, Figure~(a) and ``Head'' in Tabel~(b) indicate our experiments of introducing category priors~(CP) into the verb prediction head.
    Results from (b) indicate that taking such semantic information as the Object Query initialization~(shown as Figure~\ref{fig1-curve}.a) achieves a significant performance gain than into the prediction head~(Row.4 vs. Row2/3).
    }
    \label{fig:query_head}
    \vspace{-0.1cm}
\end{figure}

\textbf{The impacts of where to leverage the category priors.} To verify how the category-aware semantic information better promotes the HOI detection model, we design experiments to leverage category priors in another location. As Figure~\ref{fig:query_head}, the category priors are introduced in prediction heads. Before predicting the categories of interaction, we combine the visual feature and the category prior by different operations~(add and concatenate). Experimental results indicate that taking category-aware semantic information as the Object Query initialization achieves better performance than using the information as complementary features.
\par
\textbf{The impacts of different initial types.} Table~\ref{table:types of initialization} presents comparisons to different types of query initialization, including ``Zeros'', ``Random Values~(following the Uniform or Gaussian distribution)'' and ``Category-Aware Semantic Information''.
Models of the Object Query initialized with 3 different category-aware semantic information consistently achieve better performance than other initial types.

\par
\begin{figure}[t]
  \centering
    \begin{subfigure}{0.32\columnwidth}
      \centering
      \includegraphics[width=0.9\linewidth]{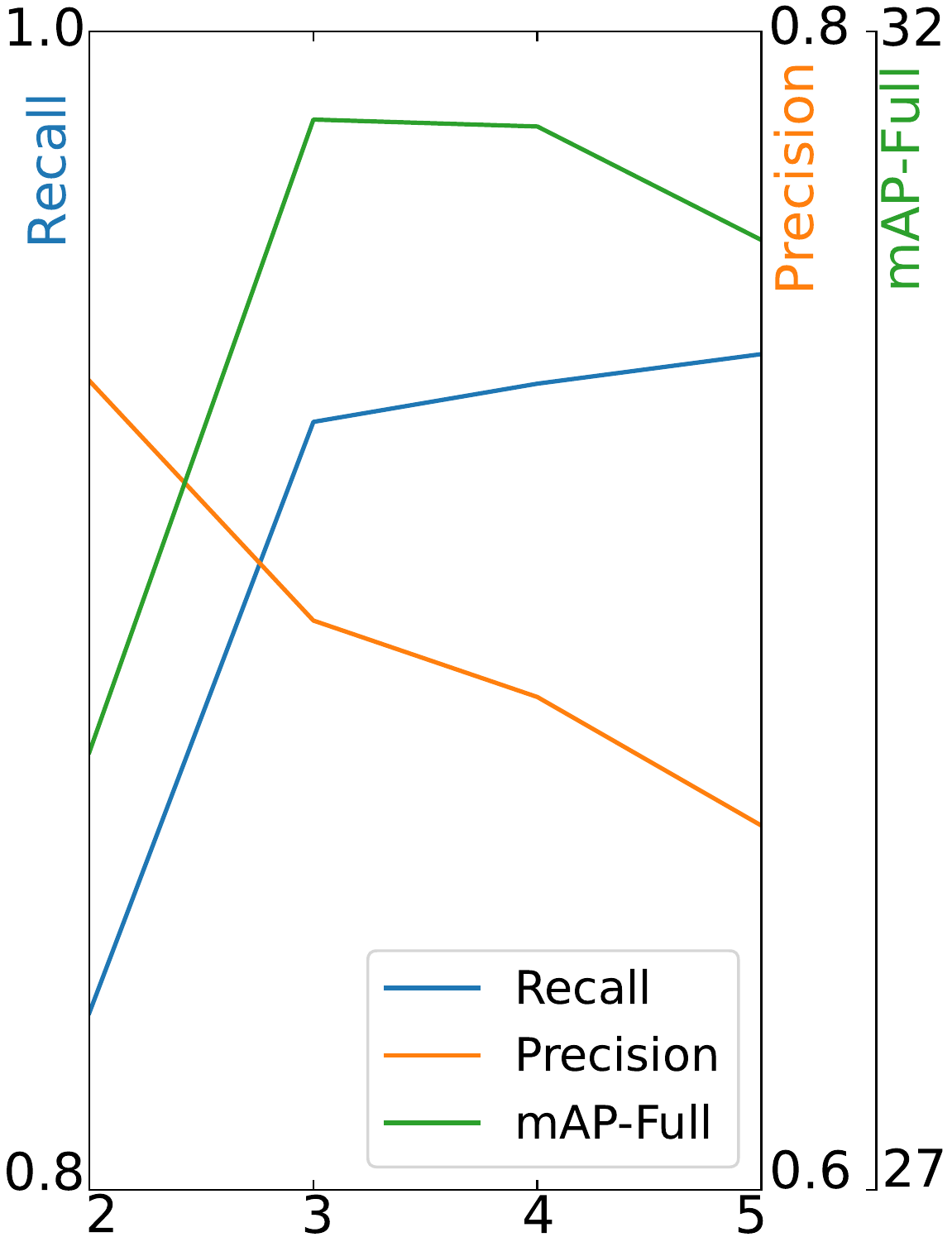}
        \caption{$N_c$.}
        \label{fig:sub1}
    \end{subfigure}   %
    \hfill  
    \begin{subfigure}{0.32\columnwidth}
      \centering
      \includegraphics[width=0.9\linewidth]{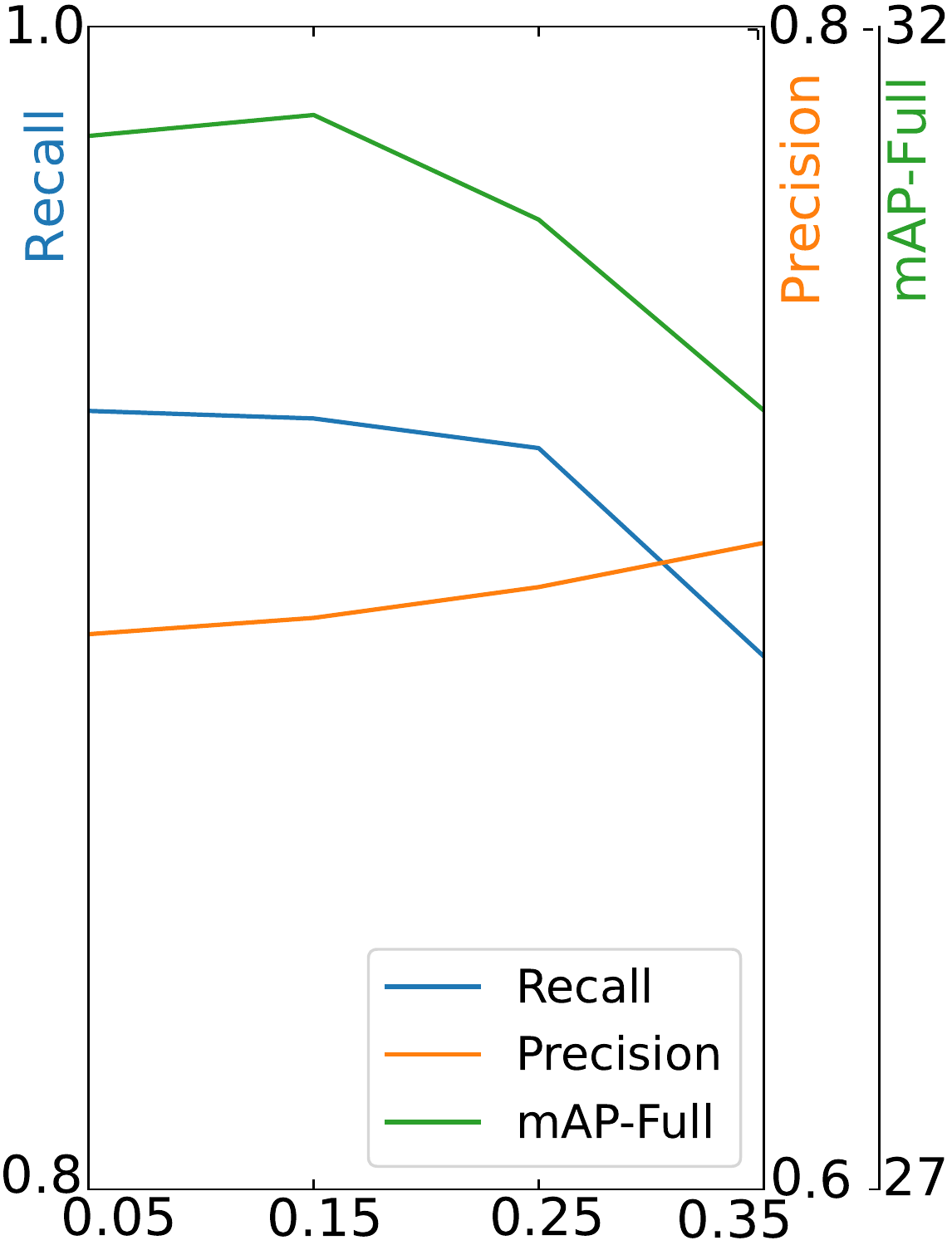}
        \caption{$T_{det}$.}
        \label{fig:sub2}
    \end{subfigure}
    \hfill
    \begin{subfigure}{0.32\columnwidth}
      \centering
      \includegraphics[width=0.9\linewidth]{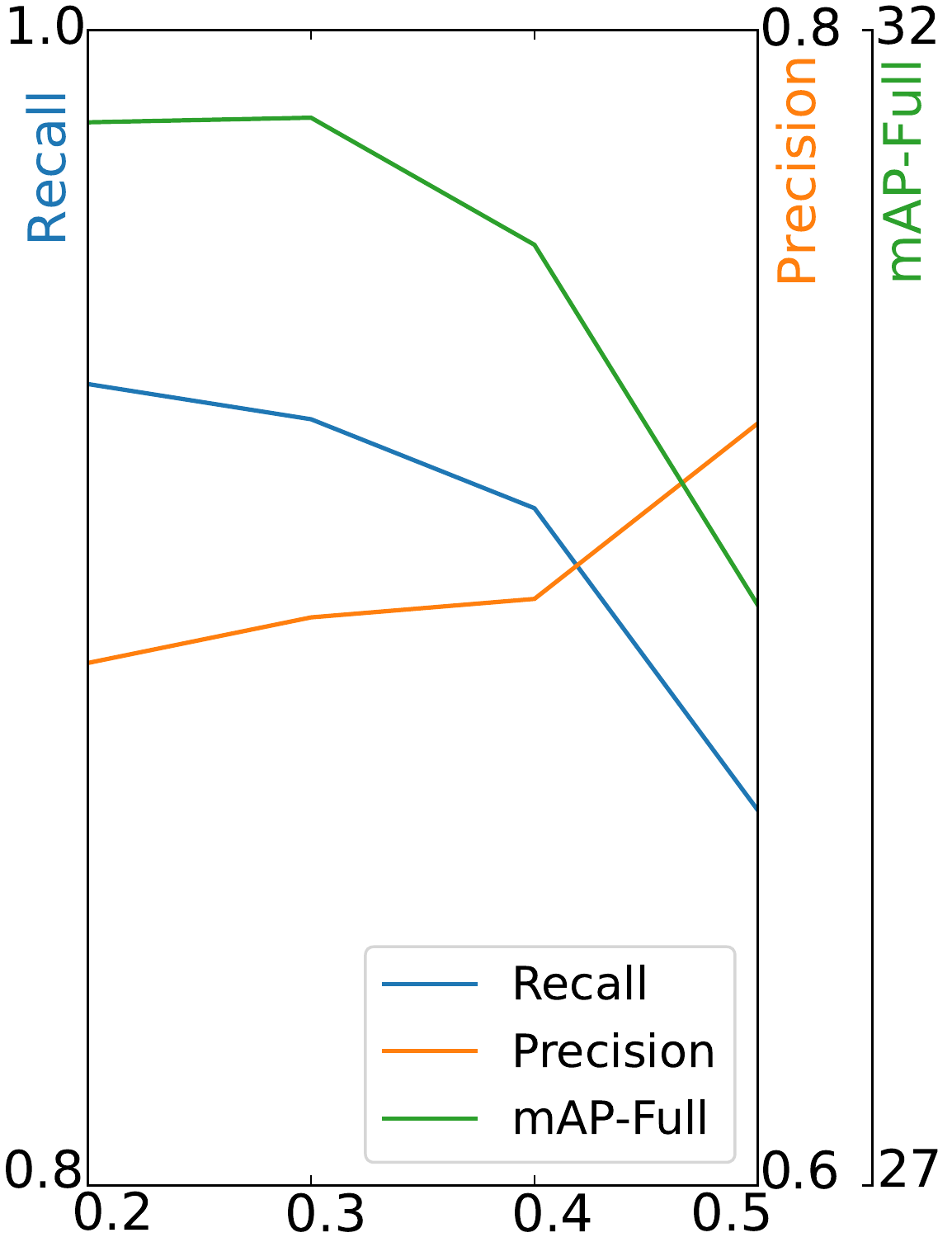}
        \caption{$T_{can}$.}
        \label{fig:sub3}
    \end{subfigure}
    \vspace{-0.2cm}
    \caption{The impacts of different settings of hyper-parameters, including $N_c$, $T_{det}$, $T_{can}$, in CAM.}
    \label{fig:hyper-parameters}
    \vspace{-0.3cm}
\end{figure}

\textbf{Hyper-parameters in CAM.} Figure~\ref{fig:hyper-parameters} illustrates the variance by several hyper-parameters, including $N_c$, $T_{det}$, and $T_{can}$, in CAM. To clearly study the impacts1 of them on the quality of category priors, we calculate the recall and precision metrics of the prior categories in image level not instance level. In other words, we only care if a object category could be detected, not the amount and location. We change one parameter in turn and keep others consistent. We achieve the best performance where $N_c=3$, $T_{det}=0.15$, and $T_{can}=0.30$, due to a better trade-off between the recall and precision.

\begin{table}[t]
    \centering
    \resizebox{0.95\columnwidth}{!}{
    \begin{tabular}{c | c | c | c c c}
        \hline
        & Method & Value & Full & Rare & Non-Rare  \\
        \hline
        1 &   Zeros   & Zero & 29.07 & 21.85 & 31.23 \\
        \hline
        2 & \multirow{2}{*}{Rondom} & Uniform & 29.70 & 23.53 & 31.53 \\
        3 &      & Gaussian & 29.60 & 22.42 & 31.73 \\
        \hline
        4 & \multirow{3}{*}{CAQ} & fastText~\cite{fastText2017mikolov-advances} & 31.03 & 23.97 & 33.12\\
        5 &         & BERT~\cite{bert2018devlin} & 31.28 & 24.89 & 33.14 \\
        6 &         & CLIP~\cite{clip2021radfordlearning} & 31.23 & 24.82 & 33.10 \\
        \hline
    \end{tabular}
    }
    \caption{The impacts of different initial types. CLAM is not used due to the need of category priors.
    Models of the Object Query initialized with 3 different category-aware semantic information consistently achieve better performance than other initial types.
    }
    \label{table:types of initialization}
    \vspace{-0.1cm}
\end{table}

\section{Conclusion}
In this paper, we explore the issue of promoting a transformer-based HOI model by initializing the Object Query with category-aware semantic information. We propose the Category-Aware Transformer Network~(CATN), which obtains two modules: CAM for generating category priors of an image and CLAM for enhancing the representation ability of the features. Extensive experiments, involving discussions of different initialization types and where to leverage the semantic information, have been conducted to demonstrate the effectiveness of the proposed idea. With the category priors, our method achieves new state-of-the-art results on both V-COCO and HICO-DET datasets.

\section{Acknowledge}
This work is supported by the National Natural Science Foundation of China~(NSFC) grant 62176100, the Central Guidance on Local Science and Technology Development Fund of Hubei Province grant 2021BEE056.

{\small
\bibliographystyle{ieee_fullname}
\bibliography{camera_read}
}

\clearpage
\section{Supplementary}
\subsection{Complexity \& Effectiveness}
Our innovative method can be regarded as a ‘Plug-and-Play’ module that could be readily implemented to effectively promote the performance of transformer-based HOI detection models.
Though the complexity of our method may slightly increase
compared to baseline models, it is controllable in practical
scenes. As shown in Table~\ref{tab:efficiency}, the inference time only increases 2.4ms per image when a lightweight detector (e.g. Yolov5m) and the parallel architecture are adopted.

\begin{table}[h]
    \centering
    \resizebox{0.95\columnwidth}{!}{
    \begin{tabular}{c|c|c|c|c}
    \hline
        Method & Detector & mAP & Inf~(ms) & Inf in Parallel~(ms) \\
    \hline
        QPIC~(baseline) & - & 29.07 & \textbf{43.7} & \textbf{43.7} \\
    \hline
        \multirow{2}{*}{CATN~(Ours)} & Faster RCNN & \textbf{31.62}  & 81.3 & 58.4~(\textcolor{black}{+14.7})\\
         & Yolov5m & \underline{31.49}  & \underline{57.2} & \underline{46.1}~(\textcolor{red}{+2.4}) \\
    \hline
    \end{tabular}
    }
    \caption{Our method could obtain significant improvement with the controllable increase of complexity. The inference time only increases 2.4ms per image on one RTX3090 when a lightweight detector, e.g. Yolov5m, is adopted in parallel.}
    \label{tab:efficiency}
\end{table}
\vspace{-0.5cm}

\subsection{Visualization}
\begin{figure}[thb]
    \centering
    \includegraphics[width=1.0\columnwidth]{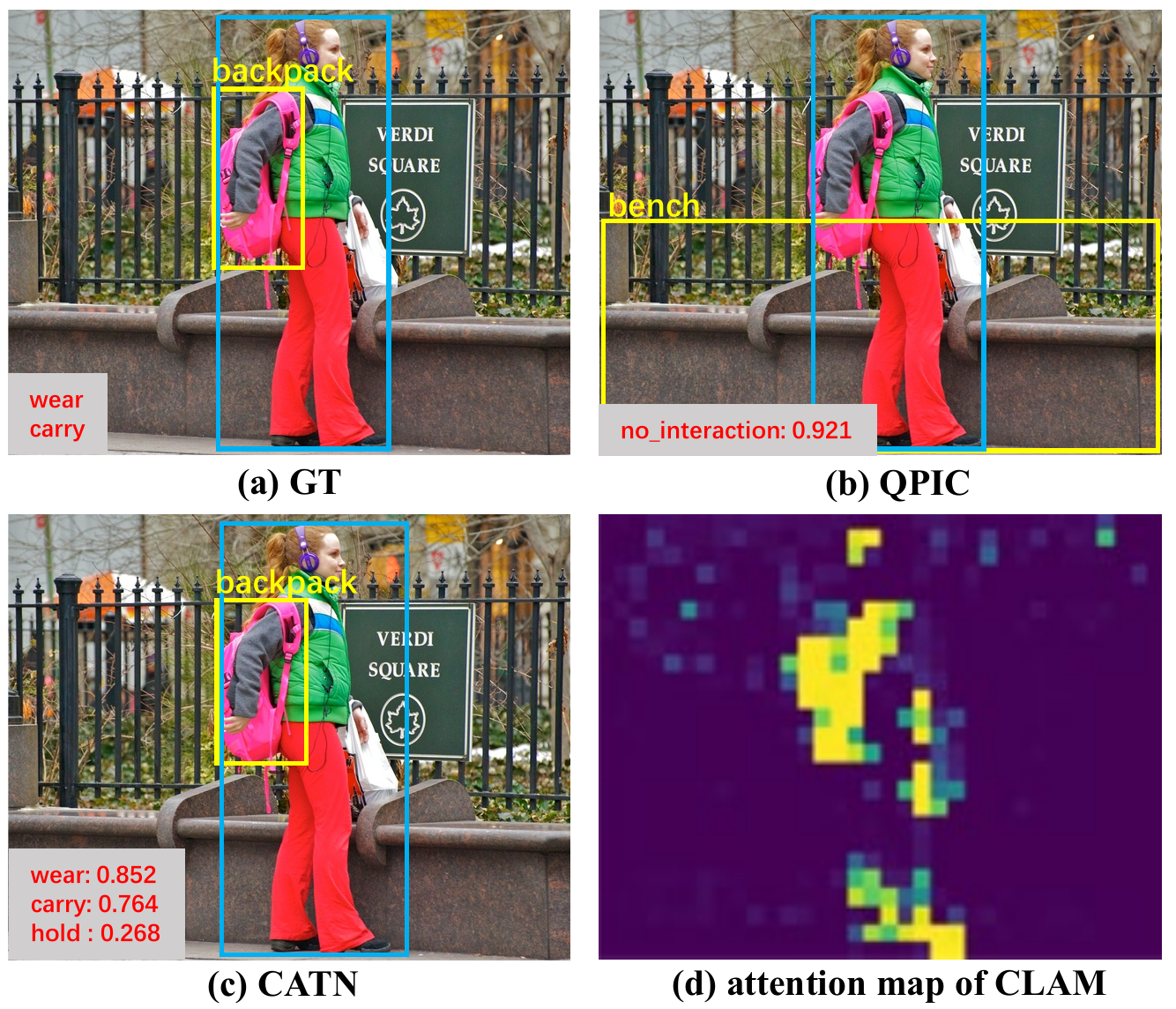}
    \caption{Visualization of a sample from HICO-DET. The human bounding boxes, object bounding boxes, object classes, and verb classes are drawn with blue boxes, yellow boxes, yellow characters, and red characters respectively.}
    \label{fig:visualization}
\end{figure}

Figure~\ref{fig:visualization} visualizes an HOI sample selected from the HICO-DET dataset.
As seen from Figure~\ref{fig:visualization}(a, b, c), our CATN outperform the baseline~\cite{qpic2021tamura} on both object detection and interaction classification. For object detection, our CATN accurately predicts the pair of human and object instances~(both categories and bounding-boxes) while there is a false positive sample generated by QPIC~($<$human, bench, no\_interaction(0.92)$>$ is not contained in the image). For interaction classification, the annotated interactions have higher confidence scores, while non-annotated interactions have lower scores.

\begin{figure}[h]
    \centering
    \includegraphics[width=1.0\columnwidth]{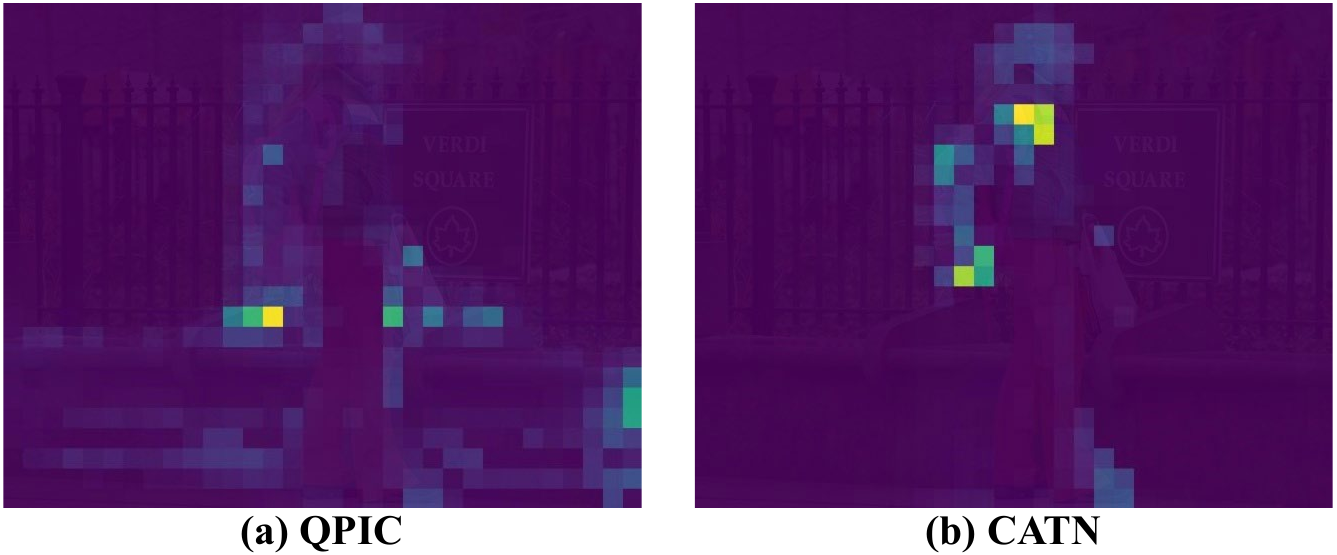}
    \caption{Visualization of decoder attention maps of QPIC~(Baseline) v.s. CATN~(Ours). As shown in Figure~\ref{fig:decoder_attention}(b), our method can lead to more accurate attention so that the aggregated features are more meaningful and more beneficial to interaction classification.}
    \label{fig:decoder_attention}
\end{figure}
\textbf{The effectiveness of CLAM.} To analyze the effectiveness of the proposed category-level attention module~(CLAM), we also
visualize the attention map of the category of `backpack' in Figure~\ref{fig:visualization}(d). The relative regions of both the `backpack' and the interacting human are highlighted in the image, which demonstrates that our CLAM can automatically aggregate the features with rich information of the corresponding category.
\par
\textbf{The effectiveness of $Q_{CA}$.} We have conducted experiments on
what initialization (all-zeros, random values, and category
information) is useful and experiments on where (Object
Query, CLAM, and Verb-Classifier) to leverage such information as shown in Tab.5 and Fig.4 in the original manuscript respectively.
This research shows that using category info to
initialize the Object Query can effectively promote the performance
of HOI detection.  To clearly explain how the category-aware information contributes to the HOI detection, we also visualize attention
maps in transformer-decoder to show that this information and initialization
would lead to more accurate attention, as shown in Figure~\ref{fig:decoder_attention}.

\subsection{HOI-NMS \& HOI-SoftNMS}
NMS~\cite{nms1971rosenfeldedge} and its variants~\cite{softnms2017bodla} can obtain the better performance when mean-Average-Precision~(mAP) is used as an evaluation metric and are therefore employed in state-of-the-art object detectors~\cite{fasterrcnn2015ren, yolov12016redmonyou, cascade2018cai}. However, theses approaches cannot be applied directly to the HOI detection since there are a pair of bounding boxes of the human and the object in one HOI prediction. Motivated by the previous~\cite{nms1971rosenfeldedge, softnms2017bodla}, we present HOI-NMS and HOI-SoftNMS methods to further improve the performance by reducing the number of duplicate HOI predictions in pair-wise level. The pseudo code and the comparison are illustrated as the following.

Based on IoU used in the object detection, we calculate $IoU_{hoi}$ in line~5 to indicate the extent of overlap between two HOI predictions where the Formula~\ref{equ:iou_hoi} shows the calculating process.

\vspace{-0.7cm}
\begin{align}
    IoU_{hoi}(h_i, h_m)=
    \begin{cases}
        $$0,\qquad\qquad If\ (c^O_i\neq c^O_m\ or\ c^I_i\neq c^I_m)$$;\\
        $$min(iou(b^H_i, b^H_m), iou(b^O_i, b^O_m)),\ Else$$;\\
    \end{cases}
    \label{equ:iou_hoi}
\end{align}

\begin{algorithm}[t]
	\renewcommand{\algorithmicrequire}{\textbf{Input:}}
	\renewcommand{\algorithmicensure}{\textbf{Output:}}
	\caption{HOI-NMS \& HOI-SoftNMS}
	\label{alg:HOI-NMS&HOI-SoftNMS}
	\begin{algorithmic}[1]
	\REQUIRE{$H={\{h_i = <b^H_i, b^O_i, c^o_i, c^I_i> | h_i\}}^{N}$, $S=\{s^I_i\}^N$ \\
	$H$ is the list of original HOI predictions;\\
	$b^H_i$, $b^O_i$ are the bounding boxes of the human and the object of the i-th HOI prediction;\\
	$c^O_i$, $c^I_i$ are the categories of the object and the interaction of the i-th HOI prediction;\\
	$s^I_i$ is the interaction score of the i-th HOI prediction.
	}
		\WHILE{$H \neq empty$}
		    \STATE $m \leftarrow argmax\ S$
		    \STATE $D\leftarrow D\cup h_m; H\leftarrow H-h_m$
		    \FOR{$h_i$ in $H$}
		        \STATE $Calculate$\ \textbf{$IoU_{hoi}(h_i, h_m)$} based on Formula~\ref{equ:iou_hoi};
		        \IF{NMS and $IoU_{hoi} > t_{iou}$}
		            \STATE $s_i = 0$
		        \ELSIF{NMS and $IoU_{hoi} > t_{iou}$}
		            \STATE $s_i = s_i \cdot e^{-\frac{IoU_{hoi}^2}{0.5}}$
		        \ELSE
		            \STATE $s_i = s_i$
		        \ENDIF
		    \ENDFOR
		\ENDWHILE
		\ENSURE  $D, S$
	\end{algorithmic}
\end{algorithm}

\vspace{-0.3cm}
\begin{table}[thb]
    \centering
    \resizebox{0.9\columnwidth}{!}{
    \begin{tabular}{c | c | c | c c c}
    \hline
        & Method & NMS & Full & Rare & Non-Rare \\
    \hline
        1 & \multirow{3}{*}{QPIC~\cite{qpic2021tamura}} & - & 29.07 & 21.85 & 31.23 \\
        2& & HOI-NMS & 29.91 & 21.77 & 32.32 \\
        3& & HOI-SoftNMS & 30.00 & 21.78 & 32.43 \\
    \hline
        4 & \multirow{3}{*}{\textbf{CATN~(Ours)}} & - & 31.62  & 24.28 & 33.79 \\
        5& & HOI-NMS & 32.38 & 25.14 & 34.52 \\
        6& & HOI-SoftNMS & \textbf{32.40} & \textbf{25.15} & \textbf{34.54} \\
    \hline
    \end{tabular}
    }
    \caption{Comparison against different NMS strategies with their own best performance. Results~(Row1/4 vs. Row2/3/5/6) indicate that the performance could be further improved when NMS strategy is employed. Compared with the original CATN with fastText~\cite{fastText2017mikolov-advances} in line~4, HOI-SoftNMS improves mAP-full from 31.62 to 32.40, which obtains the best performance.}
    \label{tab:nms}
\end{table}

\begin{table}[thb]
    \centering
    \resizebox{1.0\columnwidth}{!}{
    \begin{tabular}{c  c  c  c c c c}
    \hline
        $t_{iou}$ & 0.2 & 0.3 & 0.4 & 0.5 & 0.6 & 0.7 \\
    \hline
        HOI-NMS & 31.784 & 32.079 & 32.211 & 32.317 & \textbf{32.381} & 32.372\\
    \hline
        HOI-SoftNMS & 32.379 & 32.395 & \textbf{32.397} & 32.389 & 32.389 & 32.358 \\
    \hline
    \end{tabular}
    }
    \caption{The effect of different settings of $t_{iou}$ on the HICO-DET dataset. We conduct experiments based on the CATN with fastText~\cite{fastText2017mikolov-advances} and evaluate them on the mAP-full metric.}
    \label{tab:t_iou}
\end{table}

\end{document}